\documentclass[preprint]{elsarticle}

\usepackage{lineno,hyperref}
\usepackage{amssymb}
\usepackage{graphicx}
\usepackage{multirow}
\usepackage{caption}
\usepackage{subfigure}
\usepackage{amsmath}
\modulolinenumbers[5]
\usepackage{setspace}
\usepackage{color}

\journal{Pattern Recognition}

\begin{document}

\begin{frontmatter}

\title{Feature Flow: In-network Feature Flow Estimation for Video Object Detection}

\author[addr1]{Ruibing Jin}
\author[addr2]{Guosheng Lin\corref{CorrAuthor}}
\ead{gslin@ntu.edu.sg}
\author[addr3]{Changyun Wen\corref{CorrAuthor}}
\ead{ecywen@ntu.edu.sg}
\author[addr4]{Jianliang Wang}
\author[addr1]{Fayao Liu}

\cortext[CorrAuthor]{Corresponding author}

\address[addr1]{A*STAR, Institute for Infocomm Research, Singapore 138632}
\address[addr2]{School of Computer Science and Engineering, Nanyang Technological University (NTU), Singapore 639798}
\address[addr3]{School of Electrical and Electronic Engineering, Nanyang Technological University (NTU), Singapore 639798}
\address[addr4]{Hangzhou Innovation Institute of Beihang University, Autonomous Intelligent Systems}

\begin{abstract}
Optical flow, which expresses pixel displacement, is widely used in many computer vision tasks to provide pixel-level motion information. 
However, with the remarkable progress of the convolutional neural network, recent state-of-the-art approaches are
proposed to solve problems directly on feature-level. Since the displacement of feature vector is not consistent with the pixel displacement, a common approach is to 
forward optical flow to a neural network and fine-tune this network on the task dataset. With this method,
they expect the fine-tuned network to produce tensors encoding feature-level motion information. 
In this paper, we rethink about this de facto paradigm and analyze its drawbacks in the video object detection task.
To mitigate these issues, we propose a novel network (IFF-Net) with an \textbf{I}n-network \textbf{F}eature \textbf{F}low estimation module (IFF module) for video object detection. Without resorting to pre-training on any additional dataset, our IFF module is able to directly produce \textbf{feature flow} which indicates the feature displacement. Our IFF module consists of a shallow module, which shares the features with the detection branches. This compact design enables our IFF-Net to accurately detect objects, while maintaining a fast inference speed. Furthermore, we propose a transformation residual loss (TRL) based on \textit{self-supervision}, which further improves the performance of our IFF-Net. Our IFF-Net outperforms existing methods and achieves new state-of-the-art performance on ImageNet VID.
\end{abstract}


\end{frontmatter}


\section{Introduction}

Optical flow, which indicates the pixel displacement is very popular in many computer vision tasks \cite{zhu2017flow,zhu2017towards,zhu2017deep}. People usually apply it to extracting motion information.
Recently, video object detection has witnessed rapid progress by using convolutional neural networks (CNN) \cite{zhu2017flow,zhu2017towards,zhu2017deep}. These methods \cite{zhu2017flow,zhu2017towards,zhu2017deep} aim to leverage temporal information to enhance the feature representation at the current frame. Due to feature displacements existing across frames, they utilize optical flow (which expresses pixel displacement) generated by FlowNet \cite{dosovitskiy2015flownet} to estimate the feature displacement and align these feature maps according to the generated optical flow. After that, these aligned features are aggregated by weighted sum to produce an enhanced feature map for detection. In this paper, for express clearly, these methods \cite{zhu2017flow,zhu2017towards,zhu2017deep} are called optical flow based methods. Although good performances are achieved by these optical flow based methods, these methods are limited by several drawbacks as discussed below:

\begin{figure}[htbp]
	\begin{center}
		\includegraphics[width=\linewidth]{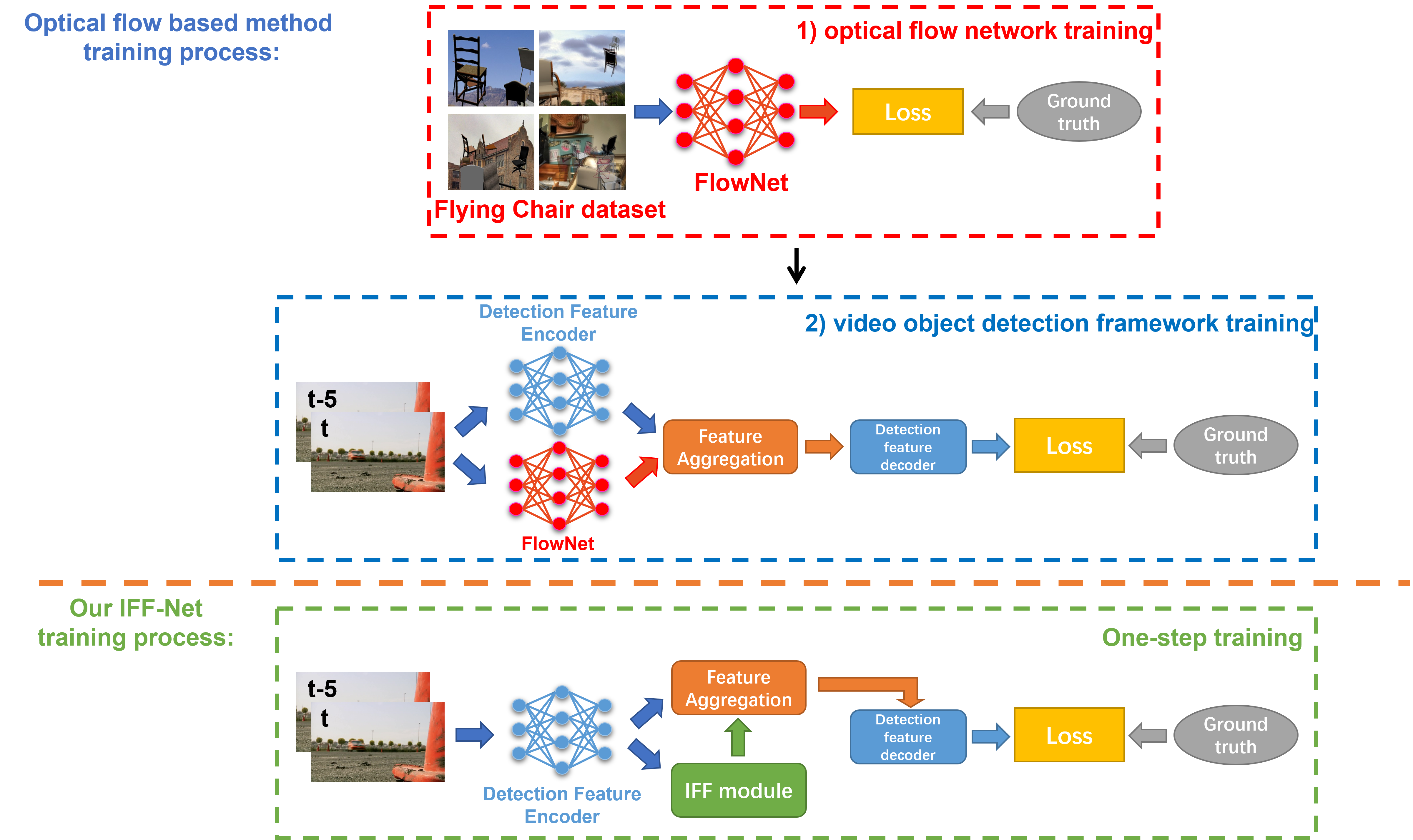}
	\end{center}
	\caption{Training process comparison between optical flow based methods and our proposed IFF-Net. The training procedure of optical flow based method is composed of two parts: optical flow network training and video object detection framework training. Different from the conventional detection method training procedure, an external training process, optical flow network training is required, where an additional dataset, Flying Chair is used to train the FlowNet to predict optical flow. After that, FlowNet is trained with other video object detection parts together on the task dataset. In comparison, our IFF-Net training process only needs one-step training, which simplifies the training process and saves more training time.}
	\label{fig:train}
\end{figure} 

\begin{itemize}
	\item \textit{Optical flow is not strictly consistent to feature displacement.} A common scheme among optical flow based methods \cite{zhu2017flow,zhu2017towards,zhu2017deep} is that they use optical flow to align features across frames. However, optical flow expresses pixel displacement in raw images. Since the feature displacement is not strictly consistent with pixel displacement, it is inaccurate to align features based on optical flow. The quality of aggregated features is degraded by this inaccurate alignment scheme (verified in Sec. \ref{exp_optflow}), limiting the accuracy of detectors. 

	\item \textit{Training procedure is complex and time-consuming due to an additionally introduced training part (optical flow network training).} Since optical flow is different from the feature displacement, it is difficult to use the original optical flow to align features across frames (discussed in Sec. \ref{exp_optflow}). To solve this problem, as shown in Fig. \ref{fig:train}, optical flow based methods \cite{zhu2017flow,zhu2017towards,zhu2017deep} firstly train an external network (FlowNet \cite{dosovitskiy2015flownet}) with an additional dataset (Flying Chair \cite{dosovitskiy2015flownet}) to predict the original optical flow. After that, to adopt the outputs from FlowNet \cite{dosovitskiy2015flownet} in feature alignment,  these optical flow based methods fine-tune the FlowNet \cite{dosovitskiy2015flownet} with detection networks on the task dataset. This makes the training procedure complex and time-consuming. 
	
	\item \textit{Inference is inefficient.} Optical flow based methods \cite{zhu2017flow,zhu2017towards,zhu2017deep} use an external network, FlowNet for feature alignment. The external network, FlowNet, consists of 23 layers, which is computationally expensive and occupies large GPU memory. This also decreases the test speed and limits the number of aggregated frames during testing.
	
	\item \textit{Domain gap exists between optical flow training datasets (Flying Chair) and the target task datasets.} Trained on an external optical flow dataset, FlowNet cannot adapt well to the video object detection task (discussed in Sec. \ref{exp_optflow}), though optical flow based methods \cite{zhu2017flow,zhu2017towards,zhu2017deep} fine-tune the FlowNet on the task dataset. This further degenerates the aggregated features and affects the accuracy of detectors.
\end{itemize} 

These optical flow based methods \cite{zhu2017flow,zhu2017towards,zhu2017deep} propose the temporal aggregation part following a well-established paradigm: pre-train on an additional dataset and fine-tune the trained model on the task dataset. However, after analyzing their drawbacks, we question this paradigm and explore a brand new method without the pre-training procedure. Our proposed method is technically different from optical flow based methods \cite{zhu2017flow,zhu2017towards,zhu2017deep} and overcomes those drawbacks in the following four aspects:

\begin{itemize}
	\item \textit{Feature flow is predicted for feature alignment, instead of optical flow.} We propose a novel module, which is able to directly predict \textbf{feature flow}, indicating the feature displacement. An \textbf{I}n-network \textbf{F}eature \textbf{F}low estimation module (IFF module) is proposed. Benefited from the feature flow generated by our IFF module, our IFF-Net can align and aggregate features better than optical flow based methods. It provides a solid foundation for accuracy improvement of detectors.
	
	\item \textit{No additional optical flow training is required.} In our proposed IFF-Net, our IFF module does not require an additionally dataset (Flying Chair) in the training procedure. Our proposed IFF module can directly learn to predict the feature flow, when it is trained with detection parts in an end-to-end manner. This simplifies the training procedure and saves much time spent on the training process.
	
	\item \textit{Our proposed IFF module shares the network backbone with detection branches.} Without resorting to an external network (FlowNet), our proposed IFF module only consists of several layers by sharing the network backbone with the detection branches. This network design avoids the additional computation overhead in FlowNet and makes feature flow estimation nearly cost-free. It also enables faster inference and less GPU memory usage. 
	
	\item \textit{Our proposed IFF module can be trained in self-supervised way.} The domain gap between the optical flow dataset and the task dataset limits the performance of the optical flow based methods. Instead of fully supervised training, we propose a transformation residual loss (TRL) to improve the performance of IFF module in \textit{self-supervised} learning. It is used as an additional loss in  the \textit{training} process. After employing TRL, feature flow from our IFF module is improved with negligibly additional computation cost. The performance of IFF-Net is also improved. 
\end{itemize}

From the four aspects above, our proposed IFF-Net mitigates the issues existing in optical flow based methods, showing a new state-of-the-art performance on the ImageNet VID benchmark \cite{russakovsky2015imagenet}. Our contributions can be summarized as follows:

\begin{itemize}
	\item 
	We propose a novel network (IFF-Net) with an in-network feature flow estimation module (IFF module). Compared with optical flow, our IFF module does not require the pre-training procedure. Without the needing of additional training datasets, the domain gap problem in optical flow based methods is avoided. Additionally, feature flow produced by our IFF module is better at feature alignment than optical flow. By sharing features with detection branches, the feature flow estimated by our IFF module is nearly cost-free, enabling efficient feature aggregation. Our IFF-Net is efficient in training and inference.
	
	\item
	IFF-Net is trained in an end-to-end manner. A transformation residual loss (TRL) is also introduced as a self-supervision, which improves our IFF module further and boosts up the performance of IFF-Net.
	
	\item Seq-NMS \cite{han2016seq} is widely used in many approaches for video object detection as a post-processing method. We improve Seq-NMS (denoted as Seq-NMS+) and apply this modified version to further boost the performance of our IFF-Net
	
	\item We have carried out extensive experiments to verify the effectiveness of our method. IFF-Net outperforms all existing methods and achieves a new state-of-the-art accuracy on the ImageNet VID \cite{russakovsky2015imagenet} while maintaining a fast inference speed.
	
\end{itemize} 

\section{Related Work}

\textbf{Object detection in static images.} Object detection methods for static images are proposed to utilize the appearance information from images to accurately detect objects. Currently, state-of-the-arts methods \cite{girshick2014rich,girshick2015fast,ren2015faster,dai2016r,lin2017feature,shrivastava2016beyond,sermanet2013overfeat,redmon2016you,liu2016ssd,lin2018focal,wang2021multi,xu2020multi,yuan2020gated,wang2018hierarchical} are mainly based on neural networks and the deep learning technology. 

These methods fall into two categories: two-stage and one-stage methods. Two-stage methods decompose object detection into two stages: proposal generation and object detection. Two-stage methods firstly utilize some algorithms to produce the proposals, which may include objects. After that, they apply a detector on these proposals to detect objects.

Among two-stage methods, R-CNN is firstly proposed in \cite{girshick2014rich}, where a CNN based method is proposed to classify each proposal region for object detection. However, this method is time-consuming and requires much disk space. To improve the training and testing speed, a ROI Pooling layer \cite{girshick2015fast} is proposed in Fast R-CNN \cite{girshick2015fast}. In Fast R-CNN, the feature map for the whole image is shared during proposal classification. By using computation sharing, the training and testing speed is improved. Then, the Region Proposal Network (RPN) in Faster R-CNN \cite{ren2015faster} is proposed by sharing the feature map of Fast R-CNN. Benefiting from the end-to-end training, Faster R-CNN shows a faster speed while remaining a better detection accuracy. Following Faster R-CNN, many detection approaches \cite{dai2016r,lin2017feature,shrivastava2016beyond} are proposed for better performance.

For one-stage methods, they detect objects only in one-step. These approaches are able to detect objects in a fast speed, while sacrificing the detection accuracy. Among deep learning based methods, OverFeat \cite{sermanet2013overfeat} is firstly proposed to use feature map from CNN to detect objects in a sliding window manner. 

After that, YOLO \cite{redmon2016you} and SSD \cite{liu2016ssd} are proposed for further improving one-stage detection methods. YOLO is short for \textit{You Only Look Once}, which is able to detect objects in a extremely fast speed, 155 fps. It unifies the detection framework into a single neural network. In YOLO, a full image is divided into a grid of sub-regions and detection results for each sub-region are predicted. Single Short multibox Detector (SSD) proposes a multi-scale feature based detection framework. Compared with YOLO, SSD progressively generates feature maps in different resolutions and applies its detector on each feature map to provide final results. These two detection methods show better results on PASCAL \cite{everingham2010pascal} than OverFeat. After that, RetinaNet \cite{lin2018focal} proposes a focal loss for improving the detection accuracy in one-stage methods and yields comparable performance to two-stage detectors.

\textbf{Object detection in video.} Different from static image based object detection, video object detection aims to utilize temporal information for performance improvement. They can be roughly divided into two types: box-level and feature-level methods. In the box level methods,  they leverage temporal information on box-level \cite{han2016seq,kang2016t,kang2016object}. The bounding boxes, which are predicted in each frame, are linked along the temporal dimension as a sequence. These linked boxes are then re-scored according to the distribution of the original scores.

Compare with box-level methods, many approaches \cite{feichtenhofer2017detect,zhu2017flow,zhu2017towards,xiao2018video,bertasius2018object,guo2019progressive,deng2019object,wu2019sequence,shvets2019leveraging} are proposed to aggregate temporal information in feature level. For the feature aggregation methods, they can be roughly divided into two categories: proposal-level and whole feature map level feature aggregation. In the proposal-level aggregation, methods in \cite{kang2017object,feichtenhofer2017detect,chen2018optimizing} train a network to predict box offsets. Approaches in \cite{wu2019sequence,shvets2019leveraging} combine the features in the related proposals based on feature similarity. These methods store proposal features in each frame and fuse them in a post-processing methods. Without the limitation of the GPU memory, these methods are able to aggregate long-range temporal information. However, limited by the proposal regions, they cannot utilize the context region to improve the detection accuracy. 

In comparison, many approaches \cite{zhu2017flow,zhu2017towards,zhu2017deep,guo2019progressive,deng2019object} are proposed to aggregate the whole feature maps across frames and detect objects on these aggregated feature maps. Due to the displacement between two frames, they firstly align feature maps across frames and fuse these feature maps together. Some methods utilize optical flow, which is produced by FlowNet\cite{dosovitskiy2015flownet}, to do feature alignment. However, optical flow is not consistent to the feature flow. Even though they finetune the FlowNet on the task dataset, their performances are not satisfactory. Additionally, FlowNet occupies much GPU memory, which causes that these methods can only aggregate nearby temporal information. This further degrades their performances. To alleviate these problems, some methods \cite{xiao2018video,guo2019progressive,bertasius2018object} try to explore temporal information on the whole feature map without FlowNet. Methods in \cite{xiao2018video,guo2019progressive} aggregate features according to the feature vector similarity. STSN \cite{bertasius2018object} directly uses deformable convolution layers to aggregate features across frames. Motion estimation and feature alignment across frames are not explicitly involved in STSN. 

Compared with the methods discussed above, we propose IFF-Net, which directly predicts feature flow and enable us to align feature map. Our IFF module shares the network backbone with detection branches. This prevents our IFF-Net from the overhead computation in FlowNet and enable our IFF module to leverage long-range temporal information. IFF-Net does not need any external dataset for training. Compared to other methods, our IFF-Net achieves better performance, though we use a smaller network and require fewer training datasets.

From the aspect of learning feature flow, TVNet \cite{fan2018end} is similar to our proposed method. However, TVNet \cite{fan2018end} is proposed for the action recognition task, which focuses on capturing the spatial-temporal features among videos. The TVNet's output does not strictly inflect the feature flow across frames. In comparison, video object detection requires to accurately align features across frames. To achieve it, our proposed IFF module aims to accurately predict the feature flow for feature alignment. Therefore, our proposed IFF-Net is different from TVNet \cite{fan2018end}. 

\begin{figure}[htbp]
	\begin{center}
		\includegraphics[width=\linewidth]{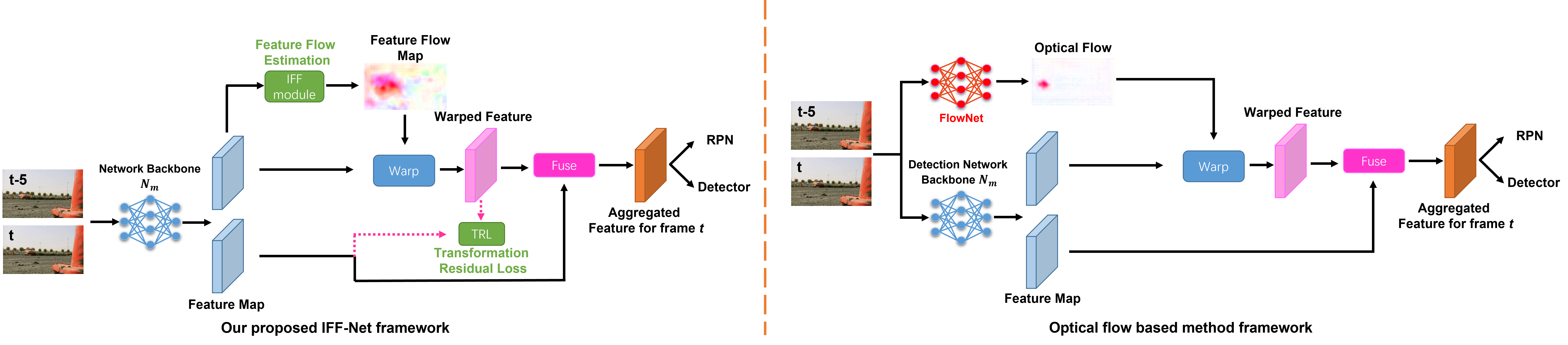}
	\end{center}
	\caption{Framework comparison between our proposed in-network feature flow estimation network (IFF-Net) and optical flow based methods. These two frameworks are illustrated in the case of fusing the features at the frame $t$ with its neighboring features at the frame $t-5$. In our proposed IFF-Net, the features at the frame $t-5$ are warped to the features at the frame $t$ via the FFM (feature flow maps) produced by our IFF module (feature flow estimation module). Then, aggregated features are generated by fusing the features at the frame $t$ and the warped features at the frame $t-5$. After that, the aggregated features are forwarded to RPN and a detector to produce results. The transformation residual loss (TRL) is applied only in the training process. In comparison, optical flow based methods consists of two neural networks. They use FlowNet to predict optical flow for feature alignment and utilizes the detection network backbone to extract feature map. This framework makes the training and testing inefficient, occupying a large amount of GPU memory.}
	\label{fig:pipeline}
\end{figure}

\section{Main Work}

Our proposed network, IFF-Net is composed of a network backbone $N_{m}$ and three branches (in-network feature flow estimation module, detector network and region proposal network), with the overall network shown in the left part of Figure \ref{fig:pipeline}. Given $L$ proceeding frames of a video, i.e. $I_{t}$, $t = 1,...,L$, $N_{m}$ produces feature maps $F_{t}$ for each of them. The number of aggregated features from neighboring frames are denoted as $\tau$. For convenience, we discuss our framework in the case of $\tau = 1$ in the following parts, as it is similar for other values of $\tau$. IFF module produces the corresponding feature flow map (FFM), $M_{j\rightarrow i}$, for features from a pair of images $I_{i}$ and $I_{j}$, where $I_{i}$ is the image at the current frame and $I_{j}$ is the image at its neighboring frame. Inspired by FGFA \cite{zhu2017flow}, the feature map from $I_{j}$ is then warped to $I_{i}$ based on the $M_{j\rightarrow i}$.  The warped feature map $F_{j\rightarrow i}$ is obtained as: 
\begin{equation}
	F_{j\rightarrow i} = \mathcal{T}(F_{j}, M_{j\rightarrow i}), 
	\label{equ:warp1}
\end{equation}where $\mathcal{T}$ is a bilinear warping function.
After that, feature aggregation is performed via a weighted sum operation on features at the current frame and other warped features according to FGFA. The aggregated feature is then passed to RPN and a detector to make predictions. As an additional loss, the TRL is used for self-supervised learning in the training process. Our IFF-Net is a single network and can be trained end-to-end. 

To distinguish our proposed IFF-Net from optical flow based methods, the framework of optical flow based methods is also illustrated in the right part of Figure \ref{fig:pipeline}. optical flow based methods utilize two neural networks: FlowNet and detection network backbone. Compared with our IFF-Net, optical flow based methods consume more GPU memory. Their training and testing are inefficient.

\subsection{In-network Feature Flow Estimation Module}

In this section, we present the in-network feature flow estimation module (IFF module) to produce \textbf{feature flow} for feature alignment across frames. Our IFF module is built in IFF-Net itself and shares network backbone with detection branches.

\begin{figure}[htbp]
	\centering
	\begin{center}
		\includegraphics[width=.5\linewidth]{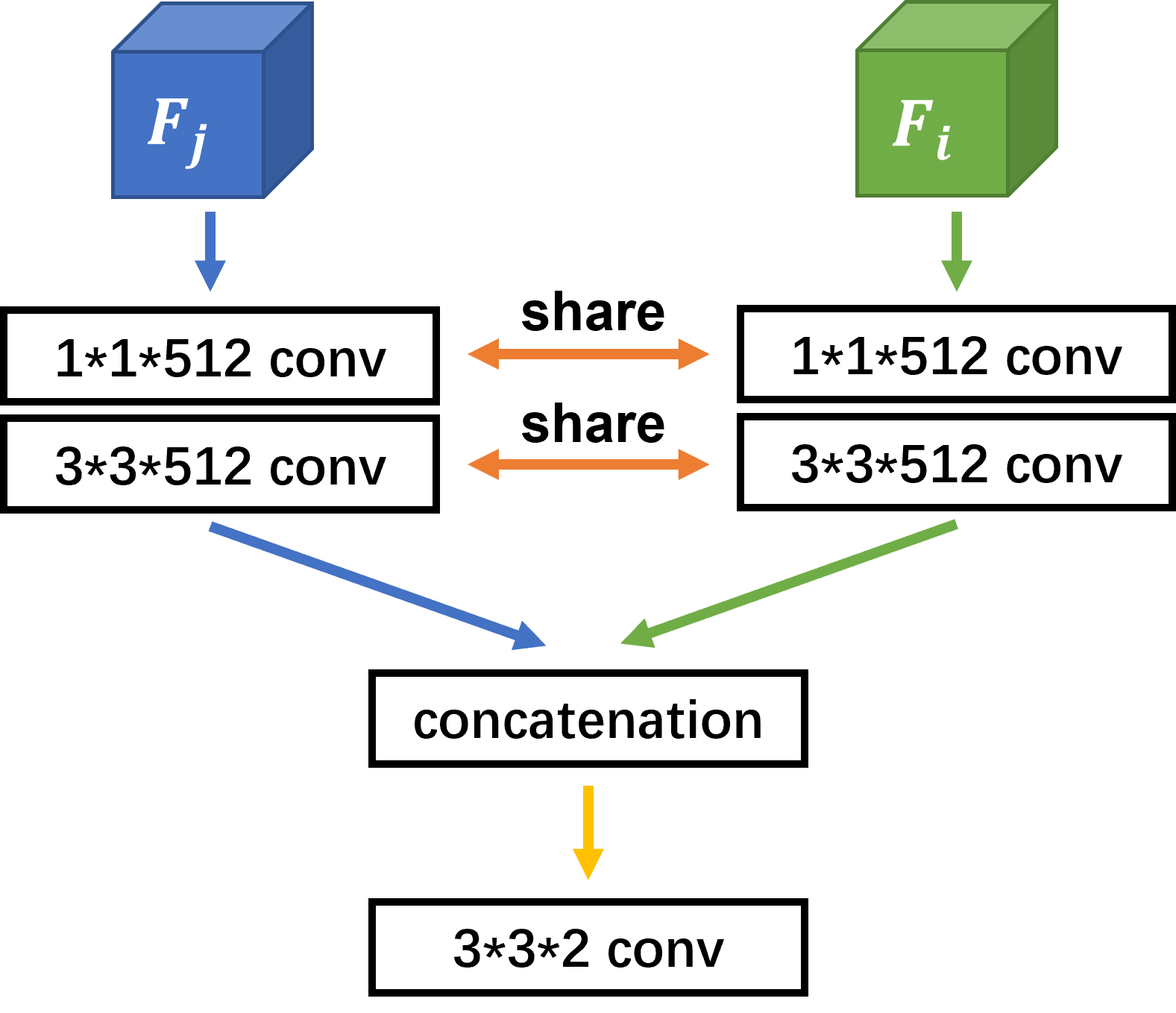}
	\end{center}
	\caption{Basic structure for feature flow estimation module (IFF module). $F_{i}$ denotes the feature of the current frame and $F_{j}$ denotes the feature of its neighboring frame. Feature $F_{i}$ and $F_{j}$ are forwarded to two convolutional layers ($1 \times 1 \times 512$ kernels and $3 \times 3 \times 512$ kernels), separately. $F_{i}$ and $F_{j}$ share the same group of these two convolutional layers. After that, they are concatenated together and pass to a convolutional layer ($3 \times 3 \times 2$ kernels) to produce the final feature flow.}
	\label{fig:fam1}
\end{figure}

\begin{figure}[htbp]
	\centering
	\begin{center}
		\includegraphics[width=.5 \linewidth]{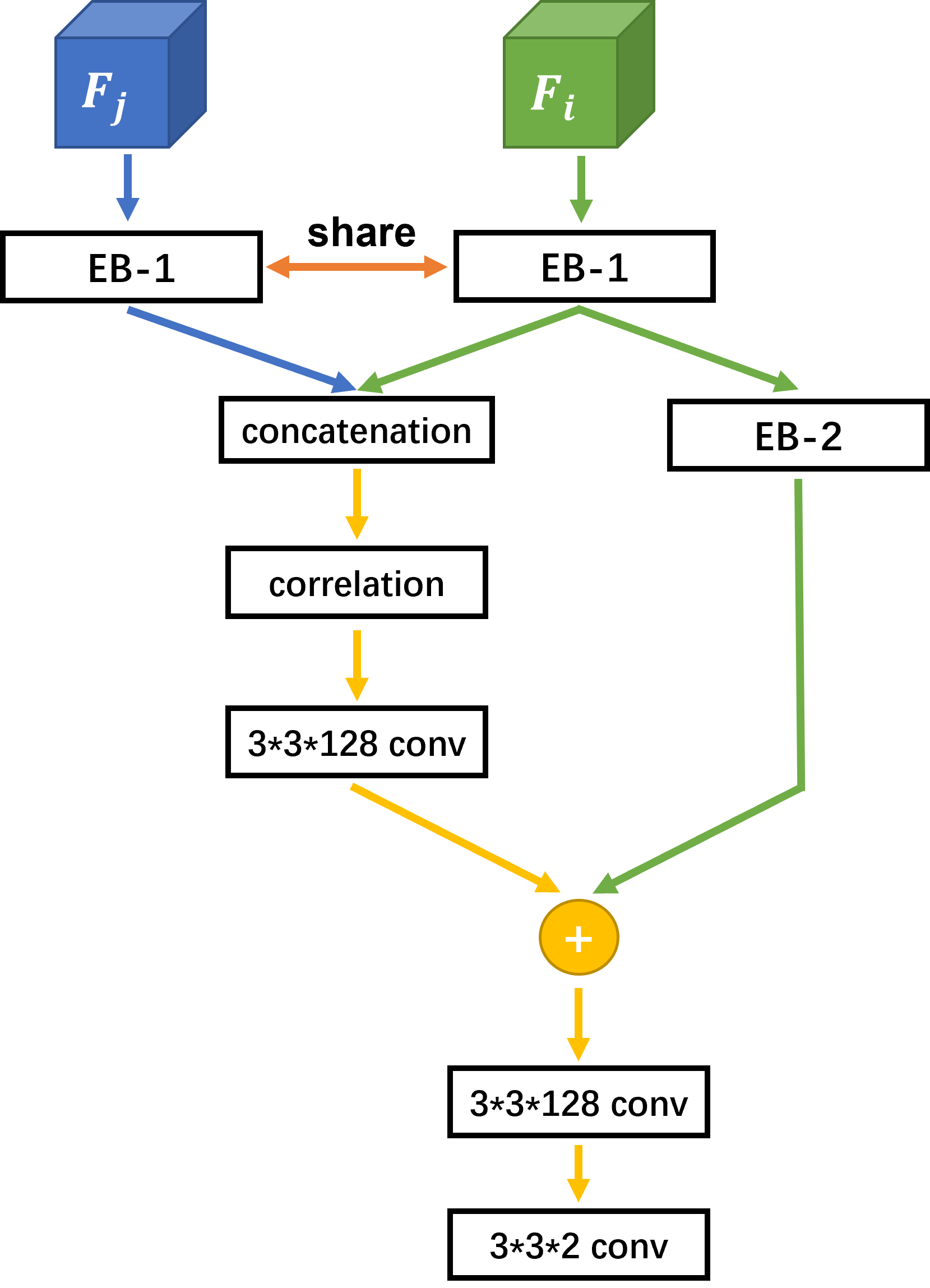}
	\end{center}
	\caption{Advanced structure for feature flow estimation module (IFF module). Feature $F_{i}$ and $F_{j}$ are forwarded to an embedded block 1 (EB-1), separately, where both EB-1 share the same parameters. Then, these two features encoded by EB-1 are concatenated together, following by a correlation layer. After that, the correlation information is converted to a 128-d feature map via a convolutional layer ($3 \times 3 \times 128$ kernels).  Simultaneously, the feature $F_{i}$ encoded by EB-1 is forwarded to an embedded box 2 (EB-2), producing a 128-d feature map. Next, an element wise addition is applied on both outputs for feature fusion. Finally, two consecutive convolutional layers are applied on the fused feature to produce the feature flow.}
	\label{fig:fam2}
\end{figure}

Given two feature maps $F_{i}$ and $F_{j}$, our IFF module predicts a feature flow map (FFM), which expresses feature flow. Two kinds of IFF modules are introduced in this part, as illustrated in Figure. \ref{fig:fam1} and Figure. \ref{fig:fam2}, respectively. 

\textbf{Basic IFF Module.} Basic IFF module is illustrated in the Figure \ref{fig:fam1}, which is a basic version of our IFF module. It consists of three convolutional layers and a concatenation layer. A feature map of the current frame $F_{i}$ and a feature map of its neighboring frame $F_{j}$ are fed into the first two convolutional layers ($1 \times 1 \times 512$ kernels and $3 \times 3 \times 512$ kernels), separately. After that, these two produced features are concatenated together and forwarded to a convolutional layer ($3 \times 3 \times 2$ kernels) which predicts the corresponding FFM. 

Benefiting from end-to-end training, our proposed IFF-Net with basic IFF module performs comparably to optical flow based methods like FGFA. FlowNet used in FGFA is composed of 23 layers, whereas our basic IFF module only consists of 4 layers. This shows that our basic IFF module is able to align features across frames in a more accurate and faster way than FlowNet. To further improve our IFF module performance, we propose an advanced version as follows.  

\textbf{Advanced IFF Module.} To improve the performance of our IFF module, an advanced IFF module is proposed in this part. The spatial relationship between the current frame and its neighboring frame may provide important information for feature flow estimation. Inspired by it, we explicitly produce this relation information and fuse it with the feature map of the current frame (which undergoes an element-wise addition merger) in the advanced IFF module. The structure of this advanced IFF module is illustrated in Figure \ref{fig:fam2}. 

\begin{figure}[htbp]
	\centering
	\begin{subfigure}[]{
			\includegraphics[width=.45\textwidth]{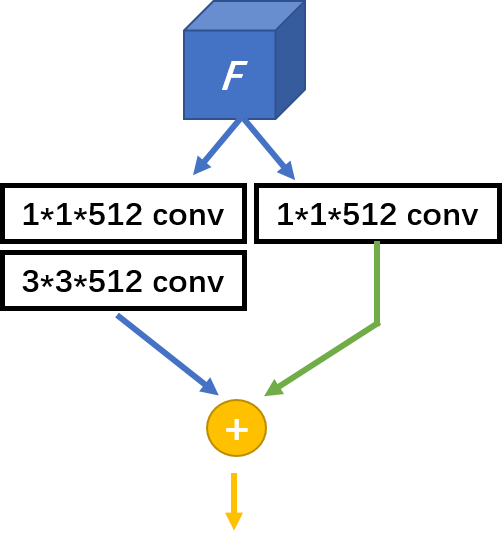}}
	\end{subfigure}
	\begin{subfigure}[]
		{\includegraphics[width=.45\textwidth]{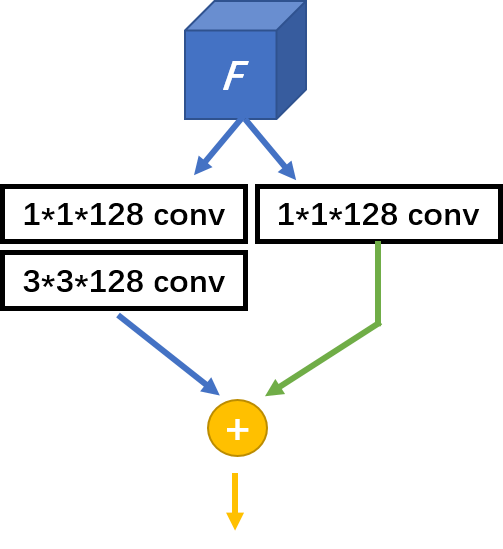}}
	\end{subfigure}
	\caption{Illustration of the structure of EB-1 (shown in (a)) and EB-2 (shown in (b)). In EB-1, feature $F$ is forwarded to two paths, respectively. The left path consists of two convolutional layers ($1 \times 1 \times 512$ kernels and $3 \times 3 \times 512$ kernels). The right path is a single convolutional layer ($1 \times 1 \times 512$ kernels). Then, outputs from both paths are fused via an element wise addition layer. The EB-2 is similar to EB-1. The difference between both blocks is that the channel number in Eb-2 decreases from 512 to 128.}
	\label{fig:2ems}
\end{figure}

Inspired by ResNet \cite{he2016deep}, the first two layers in the basic IFF module is replaced by a residual block, which is called embedded block 1 abbreviated as EB-1. Its structure is shown in Figure \ref{fig:2ems} (a).
It consists of three convolutional layers with $1 \times 1 \times 512$ kernels, $1\times1\times 512$ kernels, and $3\times 3 \times 512$ kernels.

After that, the two features are concatenated together and passed to a correlation layer \cite{dosovitskiy2015flownet} to produce the relation information between two features. The correlation layer is defined as:

\begin{align}
	o(x,y) = \frac{\textbf{F}_{i}(x, y) \cdot \textbf{F}_{j}(x + d_{x}, y+ d_{y})}{c},
	\label{equ:corr_1}
\end{align} where $\textbf{F}_{i}(x, y)$ and $\textbf{F}_{j}(x + d_{x}, y+ d_{y})$ are two feature vectors at $(x, y)$ and $(x + d_{x}, y+ d_{y})$, respectively.
$c$ is the channel number in this feature map and $\textbf{d} = (d_{x}, d_{y})$ denotes the offset between two feature vectors, $F_{i}(x, y)$ and $F_{j}(x + d_{x}, y+ d_{y})$. Correlation is operated in a square neighborhood $D \times D$ in pixel with $D = 2\bar{d}+1$, where $\bar{d}$ is the maximum displacement. A stride $s$ is used in $F_{j}$ to sample on the
feature map $F_{j}$. Therefore, the size of correlation feature map is $w\times h\times (2\times \bar{d}/s +1)^2 $, where $w$ and $h$ are the width and
height of the feature map $F_{i}$ or $F_{j}$. 

To perform element-wise addition merger, the relation information and the feature map of the current frame are converted into a 128-dimension feature map. The result of the correlation layer is passed to a convolutional layer ($3\times 3 \times 128$ kernels). The features of the current frame produced by EB-1 are passed to another embedded block denoted as EB-2, which is illustrated in Figure \ref{fig:2ems} (b). EB-2 is similar to EB-1. The difference is that the channel of kernels for layers in EB-2 are changed to 128. After merging, the fused feature map is passed to two convolutional layers ($3\times 3\times 128$ kernels and $3\times 3\times 2$ kernels) to produce a FFM. 

After applying our advanced IFF module to IFF-Net, better performance can be achieved. Compared with optical flow based methods \cite{zhu2017flow,zhu2017deep}, our IFF-Net with this advanced IFF module is significantly better, though this advanced IFF module only consists of 9 layers, which is much smaller than Flownet (23 layers). To further improve the performance of our IFF-Net, we develop a Transformation Residual Loss (TRL) and present it in the next part.

\subsection{Transformation Residual Loss}  

Since features are highly related to trained models, it is difficult to do annotations for feature flow in training datasets.
To further improve IFF module performance, we develop a transformation residual loss (TRL) as self-supervision. 
We adapt the main formulation in TV-L1 \cite{zach2007duality} into our framework. It is used as an additional loss to our IFF module in the training process. This modified formulation (TRL) is written as follows:
\begin{equation}
	\textbf{L}_{TRL} = \lambda\frac{1}{N_{p}*d}\sum_{\textbf{p} \in \Omega} {\rm smooth} \textbf{L}_{1}[\varPhi(F_{j}(\textbf{p})) - F_{i}(\textbf{p})],
	\label{equ:trl_1}
\end{equation} where \textbf{p} denotes a location on a feature map $F$, $\Omega$ denotes all locations of a feature map, $N_{p}$ denotes the number of
feature vectors in a feature map and $d$ represents the number of channels for a feature vector.  A trade-off weight 
$\lambda$ is added. The transformation $\mathcal{T}$ in Eq. (\ref{equ:warp1}) is used as $\varPhi (\cdot)$ in Eq. (\ref{equ:trl_1}). As this additional loss involves a transformation, we refer to it as transformation residual loss (TRL).

The motivation behind this Eq. (\ref{equ:trl_1}) is that an accurately aligned feature map $\varPhi(F_{j})$ from a neighboring frame should be similar to the feature map $F_{i}$ at the current frame. In Eq. (\ref{equ:trl_1}), the transformation $\mathcal{T}$ indicates an alignment transformation, which aligns the feature map $F_j$ according to our predicted feature flow. We compute the difference between $\varPhi(F_{j})$ and $F_{i}$, and use this difference as a loss for optimization. 

The temporal consistence between warped feature maps and the feature map from the current frame is involved explicitly in TRL. When training with TRL, the performance of IFF-Net is improved, surpassing methods \cite{zhu2017flow,zhu2017deep}. 

\subsection{Modified Seq-NMS}\label{seq_nms}

Seq-NMS \cite{han2016seq} has shown effectiveness in approaches \cite{zhu2017flow,xiao2018video}. It proposes a post-processing method, which produces sequences along the temporal dimension in three steps: sequence selection, sequence re-scoring and suppression. 

In this paper, we make two major modifications. Firstly, the non-maximum suppression (NMS) originally applied at the third step is moved to the first step. Through experiments, we find that this modification improves the speed of Seq-NMS while does not sacrifice the detection accuracy. 

Secondly, we jointly use average and max operation: $0.5* mean(\cdot) + 0.5*max(\cdot)$ at its second step. In the original Seq-NMS, only one operation (max or average) is applied at the second step.  Average operation suppresses outliers along temporal dimension, while max operation enhances scores of linked boxes. In our modified version, we try to combine their advantages together and modify the original operation at the second step. After employing our modifications, The performance of Seq-NMS is further improved. For convenience, this modified Seq-NMS is denoted as Seq-NMS+. 

\subsection{Network Architecture}       

The network structures in our propsed IFF-Net is presented as follows.

\textbf{Network backbone.} The ResNet-101 \cite{he2016deep} pre-trained on the ImageNet classification dataset is used as our network backbone. To meet the
requirement of our task, we modify it according to FGFA \cite{zhu2017flow}. We remove the last average pooling layer and the fully connected layers. Also in the last block, we change the effective stride from 32 to 16, and set the dilation rate of the convolutional layers as 2. 
We also add a convolutional layer ($3\times 3\times 1024$ kernels) to the last layer and initialize it with random weights.

\textbf{Detection network.} R-FCN \cite{dai2016r} is used as our detector network. We modify it according to the method in \cite{zhu2017deep}. There are two branches,
R-FCN sub-network and RPN sub-network. The first 512-dimension feature map is connected to RPN and the left 512-dimension feature map is passed to R-FCN. 9 anchors (3 scales and 3 aspects) are applied in RPN. RPN produces 300 proposals for each image. The groups in a position-sensitive score map is set as $7\times 7$. The 1024-dimension feature map of the last convolutional layer is fed into our IFF module.

\section{Experiments}

\subsection{Experimental Setting}

\textbf{Datasets.} Experiments are conducted on the ImageNet VID \cite{russakovsky2015imagenet} dataset which is for video object detection. There
are 3862 training videos and 555 validation videos with annotations of 30 classes. We train our model on a subset of ImageNet DET training set (only the same 30 classes as in ImageNet VID) and VID training set following the setting in previous methods \cite{zhu2017flow,kang2017object}. Following their protocols, training is performed on the training set and performance is evaluated on the validation set. 

\textbf{Motion Categories.} Following the protocol of FGFA \cite{zhu2017flow}, we divide objects in ImageNet VID into three categories according to object motion speed. The object motion speed is measured based on the average of object ground truth intersection-over-unions (IOUs) across frames $[t-10,t+10]$. The objects with average IOU larger than 0.9 are classified into the slow category. For objects with average IOU $\in [0.7, 0.9]$, they belong to the middle category and the remaining objects are classified into the fast category. According to FGFA, the slow, middle and fast motion categories respectively occupy $37.9\%$, $35.9\%$ and $26.2\%$.

\textbf{Implementation details.} SGD optimization is performed in the training process. 120\textbf{K} iterations are performed on 4 GPUs. Each GPU holds one image sampled from ImageNet DET or VID dataset. A learning rate of $10^{-3}$ is used for the first 71.25 $\textbf{K}$ iterations, and a learning rate of $10^{-4}$ is set for the rest iterations. Images are resized to a shorter side of 600 pixels in both training and testing. Only two neighboring frames are warped during training. For the current frame $t$, two frames are sampled randomly in the range of $[t-10,t+10]$. $\lambda$ in Eq. (\ref{equ:trl_1}) is set as 0.65. In Eq. (\ref{equ:corr_1}), the maximum displacement $\bar{d}$ is 10 and the stride $s$ is 2. In testing, the number of warped images is set as 20, which is the same as in FGFA.  

\subsection{Comparison between Optical FLow and Feature Flow}\label{exp_optflow}

\begin{figure}[htbp]
	\begin{center}
		\includegraphics[width=\textwidth]{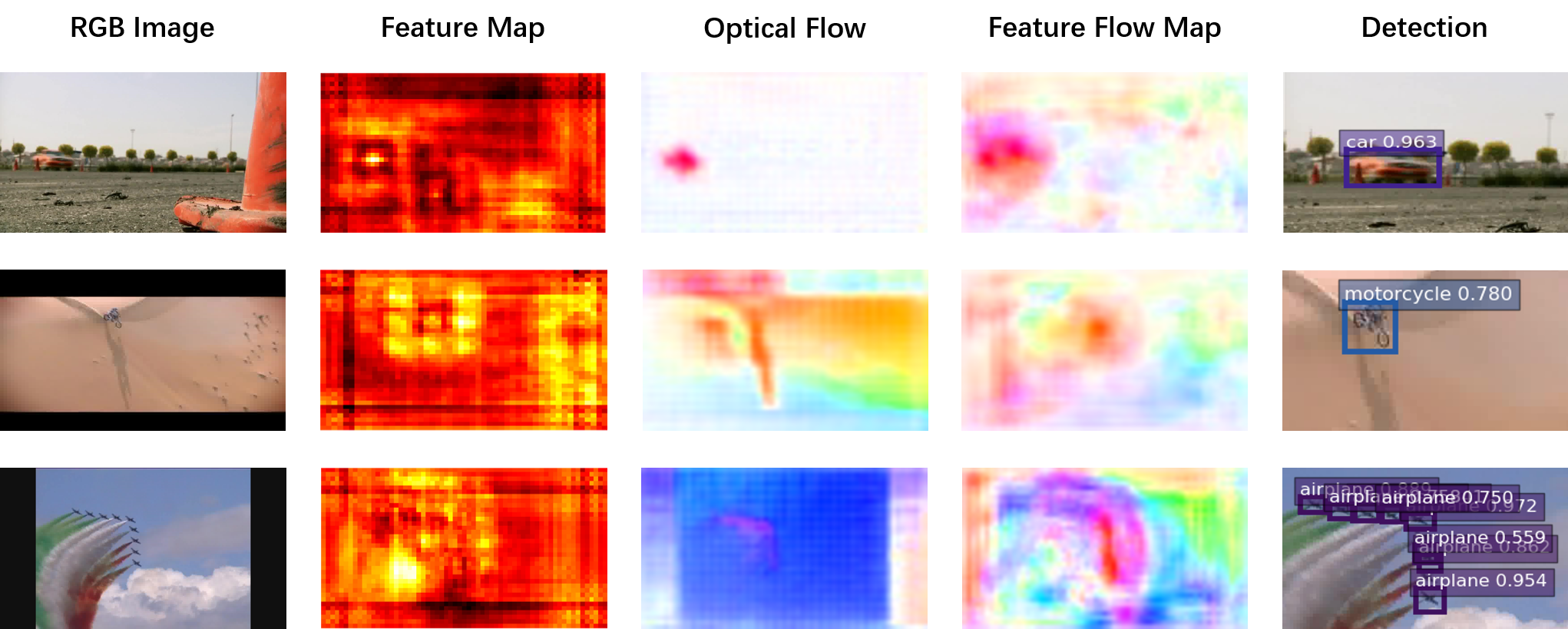}
	\end{center}
	\caption{Visualization of feature flow maps. The first to the fifth columns respectively list original images, corresponding feature map visualizations, optical flow from Flownet, feature flow maps from our IFF module and detection visualizations.}
	\label{fig:opt_flow}
\end{figure}

Optical flow indicates pixel-level displacement across frames, which is not consistent with feature-level displacement. To show the limitation of optical flow in feature alignment, we firstly conduct five experiments listed in Table \ref{table:optflow} and do comparison between optical flow and our proposed feature flow in feature alignment. Then, we illustrate some visualization results in the Figure. \ref{fig:opt_flow} and briefly discuss them.

\textbf{Feature Alignment Experiments.} In Table \ref{table:optflow}, baseline(R-FCN) serves as our baseline, where we only apply R-FCN, an image based detection method, to each frame alone. There is no temporal information involved in baseline(R-FCN). In comparison, two experiments (denoted as OptFlow-A and OptFlow-B) are carried out by using FlowNet for temporal feature aggregation and R-FCN is applied to the aggregated feature. We also conduct two experiments, where we replace FlowNet with our proposed IFF module, which is denoted as IFF-Net-B and IFF-Net-C. All these five experiments are conducted under the same hyper-parameters.  

\begin{table}[htbp]
	\caption{Comparison between optical flow and feature flow in feature alignment. OptFlow-A and OptFlow-B apply optical flow to feature warping, while IFF-Net uses feature flow in feature alignment. This shows that optical flow is not suitable for feature alignment, while our feature flow is able to effectively align temporal features.}
	\begin{center}
		\scalebox{1.0}
		{\begin{tabular}{c|c|c|c|c}
				\hline
				Method & mAP & mAP (slow) & mAP (middle) & mAP (fast) \\
				\hline
				baseline(R-FCN) &74.1 & 83.6       & 71.6         & 51.2       \\
				\hline
				OptFlow-A & 74.4 & 84.9 & 74.0 & 51.5 \\
				\hline
				OptFlow-B & 75.4 & 83.7 & 74.8 & 53.5 \\
				\hline
				\hline
				IFF-Net-B(ours) & 76.4 & 85.3 & 75.0         & 54.6       \\
				\hline
				IFF-Net-C(ours) & 77.1 & 85.0 & 75.9       & 57.7       \\ 
				\hline
		\end{tabular}}
	\end{center}
	\label{table:optflow}
\end{table}

In OptFlow-A, we use a FlowNet, which is \textbf{pre-trained} on the optical flow dataset, for feature alignment. To ensure feature alignment conducted with optical flow, we fix FlowNet and only fine-tune detection part in the task dataset. From the Table \ref{table:optflow}, it can be seen that OptFlow-A performs similar to baseline(R-FCN) (74.4 v.s. 74.1). However, in OptFlow-B, we jointly fine-tune FlowNet and the detection parts during training. Compared with OptFlow-A, OptFlow-B shows 1 mAP improvement. \textit{This shows that orignal optical flow is not suitable for feature alignment due to the difference between pixel-level displacement and feature-level displacement. }

To solve this problem, we replace the FlowNet in OptFlow-B with our proposed IFF module in IFF-Net-B and IFF-Net-C. IFF-Net-B uses our basic IFF module, and IFF-Net-C uses our advanced IFF module. From the Table \ref{table:optflow}, although our basic IFF module only consists of three layers, it effectively improve the detection accuracy than OptFlow-B (76.4 v.s. 75.4). By employing our advanced IFF module in IFF-Net-C, the performance is further improved (77.1 v.s. 75.4). 

In the experiment OptFlow-B, although we fine-tune FlowNet on the task dataset, the performance of optical flow based method (75.4) is still inferior to our IFF-Net (76.4 and 77.1). \textit{This shows that the domain gap between FlowNet training dataset and the task dataset hinders the performances of optical flow based methods.}

According to this comparison between optical flow and our feature flow, it demonstrates that optical flow is not suitable for feature alignment. Our proposed IFF modules effectively solve this problem and show significant improvement. 

\textbf{Feature Flow Visualization and Discussion.} To show our proposed feature flow clearly, we visualize some feature flow results and compare them with the corresponding optical flow in Figure. \ref{fig:opt_flow}.       

Original images are shown in the first column, while columns 2 to 4 respectively list their corresponding feature maps, the optical flows of Flownet, and the FFM produced by our IFF module. These feature maps are accumulated along the channel axis. To show the effectiveness of our feature maps, we visualize the corresponding detection results on the last column. To show them more clearly, the regions of objects are zoomed in.

In the video detection task,  we need to perform feature alignment for feature aggregation. As feature captures different information (semantics, context) from pixels, the estimated feature flow is different from optical flow (pixel displacement). Compared with optical flow, using feature flow is more suitable for performing warping in the feature level. Moreover, end-to-end training makes our estimated feature flow well adapted to our detection task, which leads to better performance than optical flow based methods.

\subsection{Ablation Study}\label{abl}

To show the effectiveness of our IFF-Net, the performance of baseline and our IFF-Net variants on ImageNet VID are listed in Table \ref{table:ablat}.

\begin{table}[htbp]
	\caption{Ablation study of our IFF-Net and comparison with optical flow based methods. It shows that IFF module, TRL and Seq-NMS+ effectively improve the performance of IFF-Net. }
	\begin{center}
		\scalebox{0.7}{
			\begin{tabular}{c|c|c|c|c|c|c|c}
				\hline
				Method & IFF module & TRL & Post-process & mAP & mAP (slow) & mAP (middle) & mAP (fast) \\
				\hline
				baseline(R-FCN) &  &  &  &74.1 & 83.6       & 71.6         & 51.2       \\
				\hline
				IFF-Net-A(ours) & basic &  &   & 76.0 & 85.0 & 74.8         & 54.7       \\
				\hline
				IFF-Net-B(ours) & basic & \checkmark &   & 76.4 & 85.3 & 75.0         & 54.6       \\
				\hline
				IFF-Net-C(ours) & advanced & \checkmark &  & 77.1 & 85.0 & 75.9       & 57.7       \\
				\hline
				IFF-Net-D(ours) & advanced & \checkmark & Seq-NMS & 78.9 & 86.6 & 78.1 &  59.8            \\
				\hline
				IFF-Net-E(ours) & advanced & \checkmark & Seq-NMS+ & \textbf{79.7} & \textbf{87.5} & \textbf{78.7}  & \textbf{60.6}     \\ 
				\hline
		\end{tabular}}
	\end{center}
	\label{table:ablat}
\end{table}

Our baseline consists of ResNet-101 \cite{he2016deep} and R-FCN \cite{dai2016r} detector. It is a single-frame framework and the same as the baseline in the Table \ref{table:optflow}. Its mAP is 74.1$\%$ on overall categories. In Table  \ref{table:ablat}, the mAP for fast motion category, 51.2\%, is the lowest among three categories. This indicates that fast motion category is challenging. However, as the percentage ($26.2\%$) of fast motion category in the validation set is the lowest in all categories, the mAP for all categories is still very high, that is $74.1\%$.

After adding the basic IFF module, the performance (denoted as IFF-Net-A) for all categories is 76.0$\%$ mAP, 1.9\% higher than our baseline. This shows that the basic IFF module improves the performance effectively. There are a $3.5\%$ gain in mAP for fast motion category, $3.2\%$ gain in mAP for middle motion category and $1.4\%$ gain in mAP for slow category, respectively. After training with TRL (denoted as IFF-Net-B), the performance of the basic IFF module improves further (76.4\% mAP vs 76.0\% mAP). According to our experiments, IFF-Net-B saves 10\% training time, while achieving similar performance to IFF-Net-A. This indicates that TRL not only improves the performance, but also makes the convergence faster. IFF-Net-B in Table \ref{table:ablat} is the same as IFF-Net-B in Table \ref{table:optflow}.   

Another variant using the advanced IFF module and TRL, denoted by IFF-Net-C, is shown in the table. The mAP score increases from $76.4\%$ to $77.1\%$. The performance of slow motion category is nearly unchanged after replacing basic IFF module with advance IFF module, but the mAP of fast motion category increases significantly by $3.1\%$ (57.7 \% mAP vs 54.6 \% mAP). This indicates that advanced IFF module focuses on the difficult category (fast category), while yielding a similar performance on the easy category (slow category). IFF-Net-C in Table \ref{table:ablat} is the same as IFF-Net-C in Table \ref{table:optflow}.    

Seq-NMS is widely used in existing methods, and we also explore its effectiveness in our framework. After adding Seq-NMS, the performance of our IFF-Net obtains 1.8$\%$ point gain. This shows that Seq-NMS is also effective for our framework. By changing the Seq-NMS into Seq-NMS+ (our proposed extension of Seq-NMS), the performance improves further, with $0.8\%$ mAP gain obtained.

\subsection{Comparison with the state-of-the-art}\label{dis_speed}

Comparisons with the state-of-the-art methods are further conducted. Existing methods utilize different post-processing methods. For a fair comparison, we firstly compare IFF-Net with some methods listed in Table \ref{table:other0} without any post-processing method. All methods listed in Table \ref{table:other0} apply the same network backbone (ResNet-101). It can be seen that our IFF-Net-C (77.1\% mAP) is better than FGFA \cite{zhu2017flow} (76.3\% mAP) and  D\&T \cite{feichtenhofer2017detect} (75.8\% mAP).

\begin{table}[htbp]
	\caption{Comparison with state-of-the-art methods without any post-processing method. It shows that our IFF-Net-C achieves the best performance among other methods without post-processing method.}
	\begin{center}
		\scalebox{1.0}{
			\begin{tabular}{c|c|c|c}
				\hline
				Method & Backbone & Post-process & mAP  \\
				\hline
				IFF-Net-C(ours) & ResNet-101 & No & 77.1 \\
				\hline
				D\&T  & ResNet-101 & No & 75.8   \\
				\hline
				FGFA  & ResNet-101 & No & 76.3    \\
				\hline
		\end{tabular}}
	\end{center}
	\label{table:other0}
\end{table}

For more comprehensive comparison, we discuss other methods in Table \ref{table:other}, where detailed configurations for each approach are also listed. Deformable convolutional layers \cite{dai2017deformable} is widely used in existing works, denoted as DCN. We also explore its effectiveness in our IFF-Net. Following DCN v1, we train our IFF-Net with 3 DCN layers (denoted as IFF-Net-F).
 We also investigate the number of warped images in training. Previously, we only use two neighboring images in training. In Table \ref{table:other}, we further evaluate our method with 4 neighboring frames warped during training. The resulting model is denoted by IFF-Net-G which improves the performance of our IFF-Net (81.2\% mAP). We also train a model with 6 DCN layers denoted as IFF-Net-H. Our single model, IFF-Net-H, achieves the best performance (82.1\% mAP), surpassing all other methods.

\begin{table*}[htbp]
	\caption{Comparison with state-of-the-art methods under \textbf{different} settings. It shows that our IFF-Net-H achieves the best performance among other methods with less training dataset and smaller network on ImageNet VID validation set.}
	\begin{center}
		\scalebox{0.8}{
			\begin{tabular}{c|c|c|c}
				\hline
				Method & Backbone & Post-process & mAP \\
				\hline
				FGFA\cite{zhu2017flow}  &ResNet-101 + FlowNet\cite{dosovitskiy2015flownet}& Seq-NMS& 78.4 \\
				\hline
				Zhu et al. 2017\cite{zhu2017towards} &ResNet-101(DCN) +FlowNet\cite{dosovitskiy2015flownet}& None  & 78.6 \\
				\hline
				STMN\cite{xiao2018video}  & ResNet-101 + DeepMask& Seq-NMS, Ensemble & 80.5 \\
				\hline
				D\&T\cite{feichtenhofer2017detect}  & ResNet-101 &Viterbi\cite{gkioxari2015finding}  & 79.8 \\
				\hline
				STSN\cite{bertasius2018object} & ResNet-101(6*DCN) & Seq-NMS & 80.4 \\
				\hline
				PSLA \cite{guo2019progressive} & ResNet-101(6*DCN) & Seq-NMS & 81.4 \\
				\hline
				Deng et al.\cite{deng2019object} & ResNet-101(6*DCN) & Seq-NMS & 81.6 \\
				\hline
				SELSA\cite{wu2019sequence} & ResNet-101 & Seq-NMS & 80.54 \\
				\hline
				RDN \cite{deng2019relation} & ResNet-101 & None & 81.8 \\
				\hline
				LSTS \cite{jiang2020learning} & ResNet-101(6*DCN) & Seq-NMS & \textbf{82.1} \\
				\hline
				Zhang et al. 2018\cite{zhang2018integrated} & ResNet-101 & Tracklet-Conditioned\cite{zhang2018integrated} & 78.1 \\ 
				\hline
				\hline
				IFF-Net-E(ours) & ResNet-101 & Seq-NMS+ & 79.7\\
				\hline
				IFF-Net-F(ours) & ResNet-101(3*DCN) & Seq-NMS+& 80.5 \\
				\hline
				IFF-Net-G(ours) & ResNet-101(3*DCN) & Seq-NMS+ & 81.2 \\
				\hline
				IFF-Net-H(ours) & ResNet-101(6*DCN) & Seq-NMS+ & \textbf{82.1} \\
				\hline
		\end{tabular}}
	\end{center}
	\label{table:other}
\end{table*}

\begin{figure*}[htbp]
	\begin{center}
		\includegraphics[width=\textwidth]{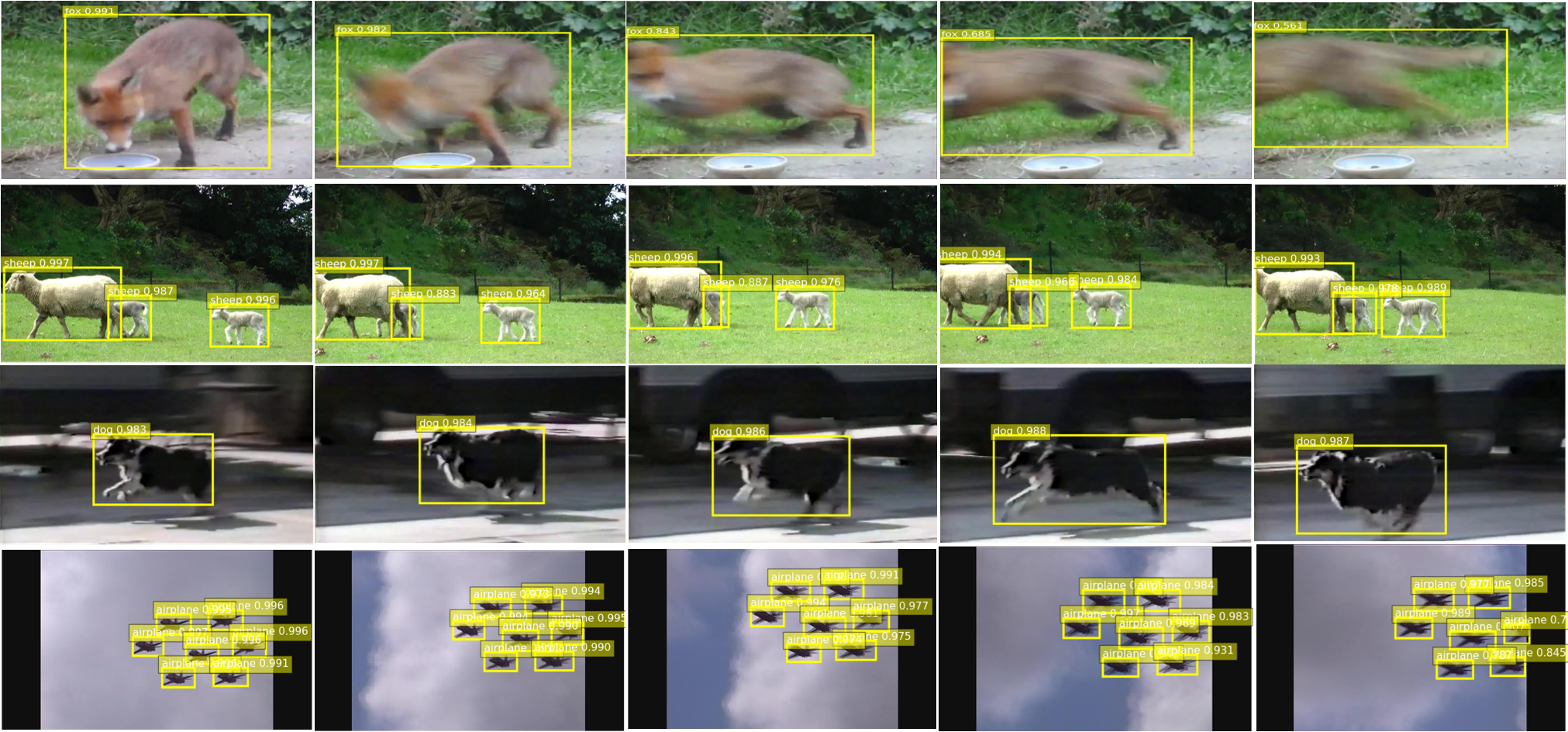}
	\end{center}
	\caption{Examples of detection results based on IFF-Net.}
	\label{fig:vis_res}
\end{figure*}

Recently, some methods uses ResNeXt as their backbone. Network backbone greatly influences the detection performance. For fair comparisons, we only report their results with ResNet-101 in this paper. Table \ref{table:other} shows that our IFF-Net-H yields the best performance among other methods. Some detection results of IFF-Net are visualized in Figure \ref{fig:vis_res}.

\begin{table}[htbp]
	\caption{Time cost comparison between FlowNet and our advanced IFF module. In general, 20 neighboring frames are aggregated in video object detection. It demonstrates that our proposed IFF module is much faster than FlowNet during temporal feature aggregation.}
	\begin{center}
		\begin{tabular}{c|c|c}
			\hline
			Method & Time per Frame & Time for 20 frames  \\
			\hline
			FlowNet & 20 ms & 400 ms \\
			\hline
			Advanced IFF module(ours)  & \textbf{8.8 ms} & \textbf{176 ms}   \\
			\hline
		\end{tabular}
	\end{center}
	\label{table:speed}
\end{table}

\textbf{Analysis on computation efficiency.} While achieving high detection accuracy, our method is also efficient. Here we analyze the time cost of our method and compare it with other existing methods in the Table \ref{table:speed}. As our IFF-Net share a similar detection pipeline with optical flow based methods, like FGFA, we only compare time cost on the feature alignment. The time cost of Flownet for optical flow prediction of one frame on our server is 20ms on average, which is similar to that reported in STMN. In comparison, our advanced IFF module only costs 8.8ms. As the number of warped images is 20 in optical flow based methods, time cost on Flownet is 400ms on average. Our IFF module only costs 176ms, which is less than half of FlowNet. In summary, our framework, IFF-Net, achieves better performance than other state-of-the-art methods in the sense that it yields a fast prediction speed with less training datasets.

\section{Conclusion}

We have proposed a novel end-to-end network, IFF-Net for video object detection.
Our IFF-Net efficiently performs feature flow estimation for effective and efficient feature alignment. Compared with other state-of-the-art methods, it requires less training datasets and uses a smaller network.  Experimental results show that our IFF-Net achieves the best results compared to other methods.

On the other hand, the predicted feature flow may not be accurate enough in the object boundary. Currently, our proposed Transformation Residual Loss (TRL) computes the differences between frames. In the future, to improve the quality of the feature flow, we will explore the context information in TRL, expecting to solve this problem. Apart from this, the existing feature fusion is designed based on the feature similarity, and the temporal offset between frames is neglected. Since temporal offset provides an important information for temporal feature fusion, we will investigate how to combine the temporal offset with the existing feature fusion model in the future.

\section*{Acknowledgments}

This work is partly supported by an NTU Start-up Grant (04INS000338C130) and MOE Tier-1 research grants: RG28/18 (S) and RG22/19 (S).

\bibliography{efan}

\end{document}